\documentclass{article}

\usepackage[final]{nips_2017}

\usepackage[utf8]{inputenc} 
\usepackage[T1]{fontenc}    
\usepackage{hyperref}       
\usepackage{url}            
\usepackage{booktabs}       
\usepackage{amsfonts}       
\usepackage{nicefrac}       
\usepackage{microtype}      
\usepackage[pdftex]{graphicx}
\usepackage{wrapfig}

\title{3D Surface-to-Structure Translation with \\ Deep Convolutional Networks}

\author{
  Takumi Moriya \\
  Keio University \\
  Kanagawa, Japan \\
  \texttt{taku3moriya@gmail.com} \\
  \And
  Kazuyuki Saito \\
  Keio University \\
  Kanagawa, Japan \\
  \texttt{saito@sfc.keio.ac.jp} \\
  \And
  Hiroya Tanaka \\
  Keio University \\
  Kanagawa, Japan \\
  \texttt{htanaka@sfc.keio.ac.jp} \\
}

\begin{document}

\maketitle

\section{Introduction}

\begin{figure}[h]
  \centering
  \label{fig:processing}
  \includegraphics[width=\linewidth]{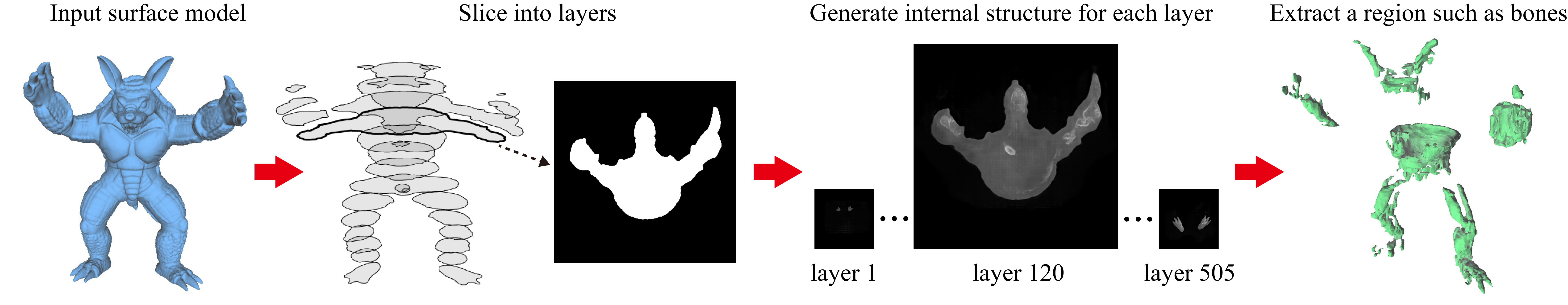}
  \caption{The process of the surface-to-structure translation. First, surface model slices into contour images, and each image is input to deep convolutional networks for the translation. Then the output images are concatenated into one volumetric data to extract a region such as bones.}
\end{figure}

Our demonstration shows a system that estimates internal body structures from 3D surface models using deep convolutional neural networks trained on CT (computed tomography) images of the human body. To take pictures of structures inside the body, we need to use a CT scanner or an MRI (Magnetic Resonance Imaging) scanner. However, assuming that the mutual information between outer shape of the body and its inner structure is not zero, we can obtain an approximate internal structure from a 3D surface model based on MRI and CT image database. This suggests that we could know where and what kind of disease a person is likely to have in his/her body simply by 3D scanning surface of the body. As a first prototype, we developed a system for estimating internal body structures from surface models based on Visible Human Project DICOM CT Datasets from the University of Iowa Magnetic Resonance Research Facility \footnote{\url{https://mri.radiology.uiowa.edu/visible_human_datasets.html}}. The estimation process given a surface model is shown in Figure \ref{fig:processing}. The input surface model is not limited to the human body. For instance, our method enables us to create Stanford Armadillo that has internal structures of the human body.

\section{Method and Experiments}

\begin{figure}[ht]
  \centering
  \label{fig:compare-images}
  \includegraphics[width=\linewidth]{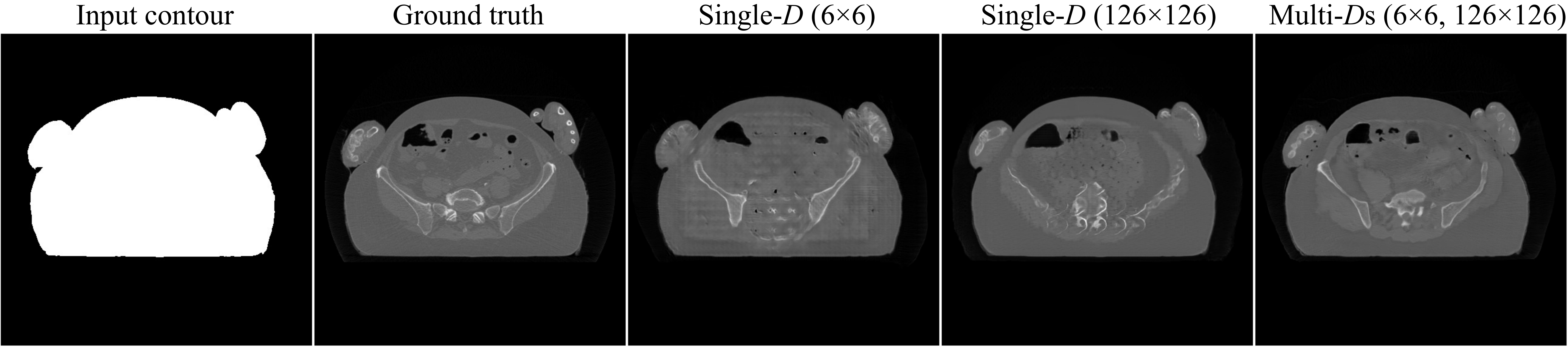}
  \caption{Comparison of the estimations on multi-$D$s and single-$D$ with various patch sizes.}
\end{figure}

When our system estimates internal structures, it uses conditional generative adversarial networks of \citet{pix2pix}. In our study, we observed that as we increase the patch size of the discriminator $D$, generated images become sharp, but the generator $G$ fails to capture the overall structures such as bones and also tiles artifacts beyond the scale. Conversely, the smaller patch size of $D$ makes images blurry, but the generator can capture the overall structures. We show this phenomenon in Figure \ref{fig:compare-images}. Therefore, we use multiple discriminators that one learns local features to sharpen images and the other learns global features to capture anatomical structures. Specifically, we construct $D$s with large patch sizes and $D$s with small patch sizes then train $G$ with these multiple discriminators $D_1, D_2, \cdots, D_N$. \citet{gman} proposed generative multi-adversarial networks that use multiple discriminators for training one generator. Thus we make use of their method. Let $x$ and $y$ be a ground truth CT image and an input contour image respectively.  $G$ is trained to minimize the following loss function:

\[
{\cal L}_G(x, y) = - \sum_{i=1}^N w_i \log(D_i(G(y))) + \lambda \| x - G(y) \|_1
\]

where $w_i$ is a weight for each discriminator $D_i$ such that $\sum_i^N w_i = 1$, and $\| \cdot \|_1$ is L1 distance. Each $D_i$ is trained independently. In our experiments, we set $\lambda$ to $100$ and used two discriminators with patch size $6\times6$ ($w = 0.25$) and $126\times126$ ($w = 0.75$). We report quantitative evaluation on various patch size in Table \ref{eval}. We compared the results by Peak Signal to Noise Ratio (PSNR) and Structural Similarity Index Measure (SSIM). Although generated samples are improved qualitatively by multi-discriminators, its metric values are almost the same as the results of a single-discriminator.

\begin{table}[ht]
  \caption{Comparison of the accuracy of the estimations on 20\% of the CT images test data.}
  \label{eval}
  \centering
  \begin{tabular}{lcccccc|c}
    \toprule
    & \multicolumn{6}{c}{Single-Discriminator} & Multi-Discriminator \\
    \midrule
    Metric    & 2$\times$2 & 6$\times$6 & 14$\times$14 & 30$\times$30 & 62$\times$62 & 126$\times$126 & (6$\times$6, 126$\times$126) \\
    \midrule
    PSNR      & 27.42 & 27.98 & 28.07 & 27.91 & 28.48 & 28.78 & 27.77 \\
    SSIM      & 0.921 & 0.926 & 0.925 & 0.929 & 0.936 & 0.938 & 0.929 \\
    \bottomrule
  \end{tabular}
\end{table}

\section{Demonstration}

\begin{wrapfigure}{r}{0.35\linewidth}
  \centering
  \label{fig:web_interface}
  \fbox{\includegraphics[width=0.9\linewidth]{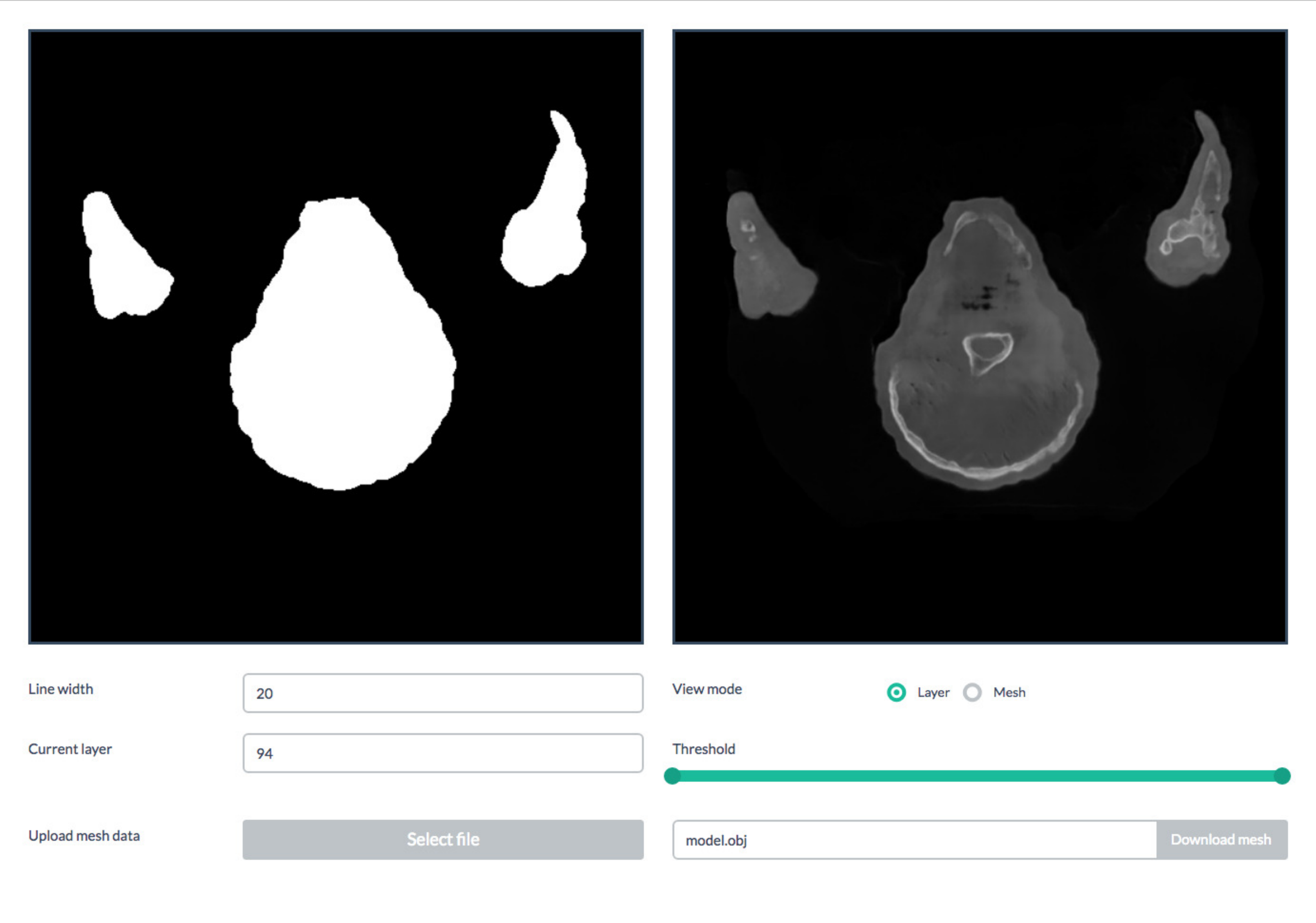}}
  \caption{Web interface.}
\end{wrapfigure}

We developed a web application to interact with the trained networks (Figure \ref{fig:web_interface}). In this web app, participants can upload 3D mesh data and obtain estimated internal structures of the data. Our demonstration consists of the following three steps. \textbf{Step 1:} Prepare a surface model by 3D scanning an object or the participant's body only when he/she is okay with being scanned. If users already have 3D models, they can skip this step. \textbf{Step 2:} Upload the data to the web app. When the upload finishes, the system estimates internal structures automatically. \textbf{Step 3:} Adjust threshold to extract a region such as bones on the web app. Finally, users can download a mesh format file of the extracted region. The last step is optional. We show some examples that created following these steps in \url{https://maxorange.github.io/internal_structure/examples.html}. For more detailed demonstration flow, please see our video here: \url{https://drive.google.com/file/d/0B02HGNE15ce4QlhCVnE5Y1BrTmc/view?usp=sharing}.

\small
\bibliographystyle{abbrvnat}
\bibliography{nips_2017}

\end{document}